\pdfoutput=1

\documentclass[11pt]{article}
\usepackage{enumitem}
\usepackage{acl}

\usepackage{times}
\usepackage{latexsym}
\usepackage{verbatim}

\usepackage[T1]{fontenc}

\usepackage[utf8]{inputenc}

\usepackage{microtype}

\usepackage[dvipsnames]{colortbl}
\usepackage{multirow}
\usepackage{graphicx}

\title{Challenges in Applying Explainability Methods to Improve the Fairness of NLP Models}

\author{Esma Balk{\i}r, Svetlana Kiritchenko, Isar Nejadgholi, and Kathleen C. Fraser \\
  National Research Council Canada \\
  Ottawa, Canada \\
 \footnotesize \texttt{\{Esma.Balkir,Svetlana.Kiritchenko,Isar.Nejadgholi,Kathleen.Fraser\}@nrc-cnrc.gc.ca}\\
 }

\begin{document}
\maketitle
\begin{abstract}
Motivations for methods in explainable artificial intelligence (XAI) often include detecting, quantifying and mitigating bias, and contributing to making machine learning models fairer. However, exactly how an XAI method can help in combating biases is often left unspecified. In this paper, we briefly review trends in explainability and fairness in NLP research, identify the current practices in which explainability methods are applied to detect and mitigate bias, and investigate the barriers preventing XAI methods from being used more widely in tackling fairness issues. 

\end{abstract}

\section{Introduction}

Trends in Natural Language Processing (NLP) mirror those in Machine Learning (ML): breakthroughs in deep neural network architectures, pre-training and fine-tuning methods, and a steady increase in the number of parameters led to impressive performance improvements for a wide variety of NLP tasks. However, these successes have been shadowed by the repeated discoveries that a high accuracy on the held-out test set does not always mean that the model is performing satisfactorily on other important criteria such as \textit{fairness}, \textit{robustness} and \textit{safety}. These discoveries that models are adversarially manipulable \citep{zhang2020adversarial}, show biases against underprivileged groups \citep{chang2019bias}, and leak sensitive user information \citep{carlini2021extracting} inspired a plethora of declarations on Responsible/Ethical AI \citep{morley2021initial}. Two of the common principles espoused in these documents are \textit{fairness} and \textit{transparency}. 

Failures in fairness of models is often attributed, among other things, to the lack of transparency of modern AI models. The implicit argument is that, if biased predictions are due to faulty reasoning learned from biased data, then we need transparency in order to detect and understand this faulty reasoning. Hence, one approach to solving these problems is to develop methods that can peek inside the black-box, provide insights into the internal workings of the model, and identify whether the model is right for the right reasons. 

As a result, ensuring the fairness of AI systems is frequently cited as one of the main motivations behind XAI research \citep{doshi2017towards, das2020opportunities, wallace2020interpreting}. However, it is not always clear how these methods can be applied in order to achieve fairer, less biased models. In this paper, we briefly summarize some XAI methods that are common in NLP research, the conceptualization, sources and metrics for unintended biases in NLP models, and some works that apply XAI methods to identify or mitigate these biases. Our review of the literature in this intersection reveals that applications of XAI methods to fairness and bias issues in NLP are surprisingly few, concentrated on a limited number of tasks, and often applied only to a few examples in order to illustrate the particular bias being studied. Based on our findings, we discuss some barriers to more widespread and effective application of XAI methods for debiasing NLP models, and some research directions to bridge the gap between these two areas.

\section{Explainable Natural Language Processing}
\label{sec:explainNLP}

\begin{table*}[t!]
\centering
\small{
\begin{tabular}{c|l|l}
\hline
 & \textbf{Local} & \textbf{Global} \\
\hline
 \multirow{5.5}{*}{\rotatebox[origin=c]{90}{\begin{tabular}{@{}c@{}}\textbf{Self-} \\ \textbf{explaining}\end{tabular}}} 
 & Gradients \cite{simonyan2013deep} &  \\
 & Integrated Gradients \cite{sundararajan2017axiomatic} & Counterfactual LM \cite{feder2021causalm} \\
 & SmoothGrad \cite{smilkov2017smoothgrad} & \\
 & DeepLIFT \cite{shrikumar2017learning} & \\
 & Attention \cite{xu2015show,choi2016retain} & \\
 & Representer Point Selection \cite{yeh2018representer} & \\
\hline
 \multirow{5.5}{*}{\rotatebox[origin=c]{90}{\textbf{Post-hoc}}} 
 & LIME \cite{ribeiro2016should} & \\
  & SHAP \cite{lundberg2017A} & TCAV \cite{kim2018interpretability,nejadgholi2022improving}\\
  & Counterfactuals \cite{wu2021polyjuice,ross2021explaining} & SEAR \cite{ribeiro2018semantically}\\
    & Extractive rationales \cite{deyoung2020eraser} & \\
  & Influence Functions \cite{koh2017understanding,han2020explaining} & \\
  & Anchors \cite{ribeiro2018anchors} & \\
\hline
\end{tabular}}
\caption{\label{tab:xai-methods}
Explainability methods from Sec.~\ref{sec:explainNLP} categorized as local vs. global and self-explaining vs. post-hoc. }
\end{table*}

With the success and widespread adaptation of black-box models for machine learning tasks, increasing research effort has been devoted to developing methods that might give human-comprehensible explanations for the behaviour of these models, helping developers and end-users to understand the reasoning behind the decisions of the model. Broadly speaking, explainability methods try to pinpoint the causes of a single prediction, a set of predictions, or all predictions of a model by identifying parts of the input, the model or the training data that have the most influence on the model outcome. 

The line dividing XAI methods, and methods that are developed more generally for understanding, analysis and evaluation  of NLP methods beyond the standard accuracy metrics is not always clear cut. Many popular approaches such as \textit{probes} \citep{hewitt2019designing,voita2020information}, \textit{contrast sets} \citep{gardner2020evaluating} and \textit{checklists} \citep{ribeiro2020beyond} share many of their core motivations with XAI methods. Here, we present some of the most prominent works in XAI, and refer the reader to the survey by  \citet{danilevsky2020survey} for a more extensive overview of the field. We consider a method as an XAI method if the authors have framed it as such in the original presentation, and do not include others in our analysis.

A common categorization of explainability methods is whether they provide \textit{local} or \textit{global} explanations, and whether they are \textit{self-explaining} or \textit{post-hoc} \citep{Guidotti2018, adadi2018peeking}. The first distinction captures whether the explanations are given for individual instances (\textit{local}) or explain the model behaviour on any input (\textit{global}). Due to the complex nature of the data and the tasks common in NLP, the bulk of the XAI methods developed for or applicable to NLP models are local rather than global \citep{danilevsky2020survey}. The second distinction is related to how the explanations are generated. In \textit{self-explaining} methods, the process of generating explanations is integrated into, or at least reliant on the internal structure of the model or the process of computing the model outcome. Because of this, self-explaining methods are often specific to the type of the model. On the other hand, \textit{post-hoc} or \textit{model-agnostic} methods only assume access to the input-output behaviour of the model, and construct explanations based on how changes to the different components of the prediction pipeline affect the outputs. 
Below, we outline some of the representative explainability methods used in NLP and categorize them along the two dimensions in Table~\ref{tab:xai-methods}.

 \textit{Feature attribution methods}, also referred to as \textit{feature importance} or \textit{saliency maps}, aim to determine the relative importance of each token in an input text for a given model prediction. The underlying assumption in each of these methods is that the more important a token is for a prediction, the more the output should change when this token is removed or changed. One way to estimate this is through the gradients of the output with respect to each input token as done by \citet{simonyan2013deep}. Other methods have been developed to address some of the issues with the original approach such as local consistency \citep{sundararajan2017axiomatic, smilkov2017smoothgrad, selvaraju2017grad, shrikumar2017learning}. 
 
 Rather than estimating the effect of perturbations through gradients, an alternative approach is to perturb the input text directly and observe its effects on the model outcome. Two of the most common methods in this class are \textit{LIME} \citep{ribeiro2016should} and \textit{SHAP} \citep{lundberg2017A}. LIME generates perturbations by dropping subsets of tokens from the input text, and then fitting a linear classifier on these local perturbations. SHAP is inspired by Shapely values from cooperative game theory, and calculates feature importance as the fair division of a ``payoff" from a game where the features cooperate to obtain the given model outcome. 
 \textit{AllenNLP Interpret} toolkit \cite{wallace2019allennlp} provides an implementation for both types of feature attribution methods, gradient based and input perturbation based, for six core NLP tasks, including text classification, masked language modeling, named entity recognition, and others.  
 
 A third way to obtain feature attribution maps in architectures that use an attention mechanism \citep{bahdanau2015neural} is to look at the relative attention scores for each token \citep{xu2015show, choi2016retain}. Whether this approach provides valid explanations has been subject to heated debate \citep{jain2019attention, wiegreffe2019attention}, however as \citet{galassi2020attention} notes, the debate has mostly been centered around the use of attention scores as local explanations. There has also been some works that use attention scores for providing global explanations based on the syntactic structures that the model attends to \citep{clark2019does}. 
 
\textit{Extractive rationales} \citep{deyoung2020eraser} are snippets of the input text that trigger the original prediction. They are similar in spirit to feature attribution methods, however in rationales the attribution is usually binary rather than a real-valued score, and continuous subsets of the text are chosen rather than each token being treated individually. Rationales can also be obtained from humans as explanations of human annotations rather than the model decisions, and used as an additional signal to guide the model. 

\textit{Counterfactual explanations} are new instances that are obtained by applying minimal changes to an input instance in order to change the model output. Counterfactuals are inspired by notions in causality, and aim to answer the question: "What would need to change for the outcome to be different?" Two examples of counterfactual explanations in NLP are \textit{Polyjuice} \citep{wu2021polyjuice} and \textit{MiCE} \citep{ross2021explaining}. Polyjuice is model agnostic, and consists of a generative model trained on existing, human generated counterfactual data sets. It also allows finer control over the types of counterfactuals by allowing the user to choose which parts of the input to perturb, and how to perturb them with control codes such as ``replace'' or ``negation''. MiCE uses model gradients to iteratively choose and mask the important tokens, and a generative model to change the chosen tokens so that the end prediction is flipped. 

There are also methods that try to pinpoint which examples in the training data have the most influence on the prediction. The most common approach for this is \textit{Influence Functions} \citep{koh2017understanding,han2020explaining}, where the goal is to efficiently estimate how much removing an example from the data set and retraining the model would change the prediction on a particular input. An alternative is \textit{Representer Point Selection} \citep{yeh2018representer}, which applies to a more limited set of architectures, and aims to express the logits of an input as a weighted sum of all the training data points. 

Some explainability methods are designed to provide global explanations using higher level, semantic concepts. \citet{feder2021causalm} use counterfactual language models to provide causal explanations based on high-level concepts. Their method contrasts  the original model representations with alternative pre-trained representations that are adversarially trained not to capture the chosen high-level concept, so that the total causal effect of the concept on the classification decisions can be estimated. \citet{nejadgholi2022improving} adapt Testing Concept Activation Vector (TCAV) method of \citet{kim2018interpretability}, originally developed for computer vision, to explain the generalization abilities of a hate speech classifier. In their approach, the concepts are defined through a small set of human chosen examples, and the method quantifies how strongly the concept is associated with a given label.

Finally, some methods produce explanations in the form of rules. One method in this category is Anchors \citep{ribeiro2018anchors}, where the model searches for a set of tokens in a particular input text that predicts the given outcome with high precision. Although Anchors is a local explainability method in that it gives explanations on individual input instances, the generated explanations are globally applicable. SEAR \cite{ribeiro2018semantically}, a global explainability method, finds universal replacement rules that, if applied to an input, adversarially change the prediction while keeping the semantics of the input the same.  

\section{Fairness and Bias in NLP Models}

Unintended biases in NLP is a complex and multifaceted issue that spans various undesirable model behaviours that cause allocational and representational harms to certain demographic groups \citep{blodgett2020language}. When the demographic group is already marginalized and underprivileged in society, biases in NLP models can further contribute to the marginalization and the unfair allocation of resources. Examples include performance disparities between standard and African American English \citep{blodgett2017racial}, stereotypical associations between gendered pronouns and occupations in coreference resolution \citep{rudinger2018gender} and machine translation \citep{stanovsky2019evaluating}, and false positives in hate speech detection on innocuous tweets mentioning demographic attributes \citep{rottger2021hatecheck}.   In this section, we review some of the most popular methods and metrics to identify such biases. For a more comprehensive coverage, see recent surveys by \citet{mehrabi2021survey} and \citet{caton2020fairness}.

Most works in ML fairness literature assume that biases in machine learning models originate from misrepresentations in training datasets and merely reflect the societal biases. However, as \citet{hooker2021moving} explains, design choices can amplify the societal biases, and automated data processing can lead to systematic un-precedented harms. \citet{shah2020predictive} identify five sources for bias in NLP models. \textit{Selection bias} and \textit{label bias} are biases that originate in the training data. The former refers to biases that are created when choosing which data points to annotate, and includes under-representation of some demographic groups as well as misrepresentation due to spurious correlations.  The latter refers to biases introduced due to the annotation process, such as when annotators are less familiar with or biased against text generated by certain groups, causing more annotation errors for some groups than others. \textit{Model bias} are biases that are due to model structure, and are responsible for the over-amplification of discrepancies that are observed in training data. \textit{Semantic bias} refers to biases introduced from the pre-trained representations, and include representational harms such as stereotypical associations. Finally, \textit{bias in research design} covers the larger issues of uneven allocation of research efforts across different groups, dialects, languages and geographic areas.

Research in fair ML has developed a number of metrics to quantify the biases in an ML model. These metrics are usually classified as \textit{group fairness metrics} and \textit{individual fairness metrics} \citep{castelnovo2021zoo,czarnowska2021quantifying}. Group fairness metrics focus on quantifying the performance disparity between different demographic groups. Some examples are \textit{demographic parity}, which measures the difference in the positive prediction rates across groups, \textit{predictive parity}, which measures the difference in precision across groups, and \textit{equality of odds}, which measures the differences between false positive and false negative rates across groups. Individual fairness metrics are based on the idea that the model should behave the same for similar examples regardless of the value of a protected attribute. A refinement to this approach is \textit{counterfactual fairness}, where the criteria for fairness is that the model decision remains the same for a given individual in a counterfactual world where that individual belonged to a different demographic group. In NLP, this notion often appears as \textit{counterfactual token fairness} \citep{garg2019counterfactual}, and is operationalized through test suites that include variations of the same text where some tokens associated with certain social groups are replaced with others, and the bias of the model is measured by the performance disparity between the pairs \citep{kiritchenko2018examining,prabhakaran2019perturbation}.   

Both group fairness metrics and individual fairness metrics are instances of \textit{outcome fairness}: whether a model is fair is determined solely on the outcomes with respect to various groups, regardless of how the algorithm produced those observed outcomes.\footnote{\textit{Outcome fairness} is also referred to as \textit{distributive fairness} in this literature.}  There is a complementary notion called \textit{procedural fairness} that is often considered in organizational settings \cite{blader2003constitutes}, which aims to capture whether the \textit{processes} that were followed to obtain the outcome are fair. In ML, this translates to whether the model's internal reasoning process is fair to different groups or individuals \citep{grgic2018beyond,morse2021ends}.
For example, outcome fairness for a resume sorting system might be implemented as ensuring that the model has the same acceptance rates or the same precision and recall for groups defined by race, gender, or other demographic attributes. A procedural fairness approach, on the other hand, might aim to ensure that the decision making process of the system only relies on skill-related features, and not features that are strongly associated with demographic attributes, such as names and pronouns.
The distinction between procedural and outcome fairness relates to different kinds of discrimination outlined in anti-discrimination laws, namely \textit{disparate treatment} and \textit{disparate impact} \citep{barocas2016big}. 

Fairness metrics have originally been developed for applications where the social group membership is known, for example in healthcare related tasks. An issue with applying these to NLP tasks is that either the demographic information is not available and needs to be estimated, or some auxiliary signal, such as the mention of a target group or the gender of the pronoun, needs to be used. However, inferring people's social attributes from their data raises important ethical concerns in terms of privacy violations, lack of meaningful consent, and intersectional invisibility \citep{mohammad2021ethics}.   Since determining whether the text is \textit{about} a certain identity group is easier than whether it is \textit{produced by} a certain identity group, there are more works investigating the former than the latter. An exception to this is the studies on disparate performance of models on certain dialects such as African American English (AAE) \citep{sap2019risk, blodgett2017racial}. This is possible due to the existence of a dialect identification tool for AAE, which was trained by pairing geo-located tweets with US census data on race \citep{blodgett2016demographic}.   

One source of bias that the NLP community has devoted significant research effort to is word embeddings and pre-trained language models \cite{bolukbasi2016man,zhao2019gender}, which \citet{shah2020predictive} characterizes as \textit{semantic bias}. Although it is not framed as such, this can be seen as a particular global explanation for biases that the models demonstrate in downstream tasks. However, the effectiveness of these methods has recently been questioned by \citet{goldfarb2021intrinsic} who found that there is no correlation between intrinsic bias metrics obtained by embedding association tests, and extrinsic bias metrics on downstream tasks.   

\section{Applications of XAI in Fair NLP}
\label{sec:XAI-FairNLP}

\begin{table*}[t!]
\centering
\small{
\begin{tabular}{ p{2.7cm}p{4cm}p{2.7cm}p{2.3cm}p{2cm}  }
\hline
\textbf{Study} & \textbf{Overall Objective of the Study} &\textbf{Application} & \textbf{Bias Type} & \textbf{Explainability Method}\\
\hline
\citet{mosca2021understanding} & Detecting classifier sensitivity towards identity terms vs. user tweet history & hate speech detection & social group bias & SHAP  \\
\citet{wich2020impact} & Measuring the effect of bias on classification performance & hate speech detection & political orientation & SHAP \\
\citet{aksenov2021fine} & Classification of political bias in news & hate speech detection & political orientation & aggregated attention scores \\ \hline
\citet{kennedy2020contextualizing} & Reducing the classifier's  oversensitivity to identity terms& hate speech detection & social group bias & feature importance (SOC)\\
\citet{mathew2021hatexplain} & Improving group fairness& hate speech detection & social group bias & LIME,  attention\\
\citet{prabhakaran2019perturbation} & Detecting biases related to named entities & sentiment analysis, toxicity detection & sensitivity to named entities & perturbation analysis\\
\citet{balkir2022necessity} & Detecting over- and under-sensitivity to identity tokens & hate speech and abusive language detection & social group bias & necessity and sufficiency\\
\hline
\end{tabular}}
\caption{\label{tab:xai-fairness-methods}
Summary of the studies that apply explainability techniques to uncover unintended biases in NLP systems. 
}
\end{table*}

To determine the uses of explainability methods in fair NLP, we search the ACL Anthology for papers that cite the explainability methods listed in Section~\ref{sec:explainNLP}, and that include keywords, ``fair'', ``fairness'', or ``bias''. We further exclude the papers that focus on other types of biases such as inductive bias, or bias terms in the description of the architecture. Our results show that although there are a number of papers that mention unintended or societal biases as wider motivations to contextualize the work (e.g., by \citet{zylberajch2021hildif}), only a handful of them apply explainability methods to uncover or investigate biases. All of the works we identify in this category use feature attribution methods, and except that of \citet{aksenov2021fine}, employ them for demonstration purposes on a few examples. 
Although our methodology excludes works that are published in venues other than ACL conferences and workshops, we believe that it gives a good indication of the status of XAI in fairness and bias research in NLP.


\citet{mosca2021understanding} use SHAP to demonstrate that adding user features to a hate speech detection model reduces biases that are due to spurious correlations in text, but introduces other biases based on user information. \citet{wich2020impact} also apply SHAP to two example inputs in order to illustrate the political bias of a hate speech model. 
\citet{aksenov2021fine} aggregate attention scores from BERT into global explanations in order to identify which words are most indicative of political bias. 


Some works beyond the papers that our search methodology uncovered on the intersection of fairness for NLP and XAI are that of \citet{kennedy2020contextualizing}, which uses Sampling and Occlusion algorithm of \citet{jin2019towards} to detect bias toward identity terms in hate speech classifiers, and that of \citet{mathew2021hatexplain}, which shows that using human rationales as an additional signal in training hate speech detection models reduces the bias of the model towards target communities. \citet{prabhakaran2019perturbation} target individual fairness, and develop a framework to evaluate model bias against particular named entities with a perturbation based analysis. Although they do not frame their model as such, the automatically generated perturbations can be categorized as counterfactuals. \citet{balkir2022necessity} suggest the use of two metrics---necessity and sufficiency---as feature attribution scores, and apply their method to uncover different kinds of bias against protected group tokens in hate speech and abusive language detection models.

As summarized in Table~\ref{tab:xai-fairness-methods}, almost all these works focus exclusively on hate speech detection, and use local feature attribution methods. The range of bias types is also quite limited. This demonstrates the very narrow context in which explainability has been linked to fairness in NLP.

There are also some works beyond NLP that use XAI to improve fairness of ML models. \citet{zhang2018fairness}, \citet{parafita2021deep}  and \citet{grabowicz2022marrying} leverage methods from causal inference to both model the causes of the given prediction and provide explanations, and to ensure that protected attributes are not influencing the model decisions through unacceptable causal chains. The disadvantage of these models is that they require an explicit model of the causal relations between features, which is a difficult task for textual data \citep{feder2021causal}. \citet{pradhan2021interpretable} also suggest a causality inspired method that identifies subsets of data responsible for particular biases of the model. \citet{begley2020explainability} extend Shapely values to attribute the overall unfairness of an algorithm to individual input features. The main limitation of all these methods is that they are currently only applicable to low dimensional tabular data. How to extend these methods to explain the unfairness of NLP models remains an open research problem.  

As abstract frameworks for connecting XAI to fair ML, \citet{deepak2021fairness} outline 
potential synergies between the two research areas.  \citet{alikhademi2021can} 
enumerate different sources of bias, and discuss how XAI methods can help identify and mitigate these.


\section{XAI for Fair NLP through Causality and Robustness}
\label{sec:causality}

The framework of \textit{causality} \cite{pearl2009causality} is invoked both in fairness and explainability literature. The promise of causality is that it goes beyond correlations, and characterizes the causes behind observations. This is relevant to conceptualizing fairness since, as \citet{loftus2018causal} argue, there are situations that are intuitively different from a fairness point of view, but that purely observational criteria cannot distinguish.

Causality tries to capture the notion of causes of an outcome in terms of hypothetical interventions: if something is a true cause of a given outcome, then intervening on this variable will change the outcome. This notion of intervention is useful for both detecting biases and for choosing mitigation strategies. Causal interventions are also the fundamental notion behind counterfactual examples in XAI. It is easier for humans to identify the cause of a prediction if they are shown minimally different instances that result in opposite predictions. Hence, causal explanations can serve as proofs of bias or other undesirable correlations to developers and to end-users.     

Going beyond correlations in data and capturing causal relations is also an effective way to increase \textit{robustness} and \textit{generalization} in machine learning models. As \citet{kaushik2020explaining} argue, causal correlations are invariant to differing data distributions, while non-causal correlations are much more context and dataset specific. Hence, models that can differentiate between the two and rely solely on casual correlations while ignoring the non-causal ones will perform well beyond the strict i.i.d. setting. 

Non-causal, surface level correlations are often referred to as \textit{spurious correlations}, and a common use case of XAI methods for developers is to facilitate the identification of such patterns. A common motivating argument in XAI methods for debugging NLP models \citep{lertvittayakumjorn2021explanation, zylberajch2021hildif}, as well as counterfactual data augmentation methods \citep{kaushik2020explaining, balashankar2021can, yang2021exploring}, is that unintended biases are due to the model picking up such spurious associations, and XAI methods which can be used to improve the robustness of a model against these spurious patterns will also improve the fairness of a model as a side effect. There is indeed evidence that methods for robustness also reduce unintended bias in NLP models \cite{adragna2020fairness,pruksachatkun2021does}. However, these methods are limited in that they can address unintended biases only insofar as the biases are present and identifiable as token-level spurious correlations. 


\section{Challenges and Future Directions}

As we saw in Sec.~\ref{sec:XAI-FairNLP} and \ref{sec:causality}, only a few studies to date have attempted to apply explainability techniques in order to uncover biases in NLP systems, to a limited extent. In this section, we discuss some possible reasons for a seeming lack of progress in this area and outline promising directions for future research. 

\paragraph{Local explainability methods rely on the user to identify examples that might reveal bias.} One issue in preventing wider adoption of XAI methods in fair NLP stems from the local nature of most explanation methods applicable to NLP models. An important step in identifying fairness problems within a model is identifying the data points where these issues might manifest. Since local explainability methods give explanations on particular data points, it is left to the user how to pick the instances to examine. This necessitates the user to first decide what biases to search for before employing XAI methods, limiting their usefulness for identifying unknown biases. 

\paragraph{Local explanations are not easily generalizable.} Even if an issue can be identified with a local XAI method, it is difficult to know to what extent the insight can be generalized. This is an issue because it is often essential to know what subsets of the input are affected by the identified biased behaviour in order to apply effective mitigation strategies. Some methods such as Anchors mitigate this problem by specifying the set of examples an explanation applies to. Other approaches use abstractions such as high-level concepts \citep{feder2021causalm, nejadgholi2022improving} to provide more generalizable insights. Principled methods to aggregate local explanations into more global and actionable insights are needed to make local explainability methods better suited to identifying and mitigating unintended biases in NLP models. Also, future NLP research could explore global explainability methods that have been used to uncover unknown biases \citep{tan2018distill}. 

\paragraph{Not all undesirable biases are surface-level or non-causal.} In the motivation for XAI methods, there is strong emphasis on identifying token-level correlations caused by sampling bias or label bias. Although methods that target these patterns are shown to also improve the fairness of models, not all sources of bias fit well into this characterization \citep{hooker2021moving}, and hence might be difficult to detect with XAI methods that provide token-level explanations. For example, \citet{bagdasaryan2019differential} show that the cost of differential privacy methods in decreasing the accuracy of deep learning NLP models, is much higher for underrepresented subgroups. A rigorous study of a model's structure and training process is required to discover such bias sources.  

Another issue that is common in works that approach fairness through robustness is the characterization of unintended biases as non-causal associations in data \citep{kaushik2020explaining, adragna2020fairness}. 
In fact, it can be argued that many of the undesirable correlations observed in data are causal in nature, and will likely hold in a wide variety of different data distributions. For example, correlations between different genders and occupations---which arguably is the source of the occupational gender stereotypes picked up by NLP models \citep{rudinger2018gender}---are not due to unrepresentative samples or random correlations in the data, but rather underlying systemic biases in the distribution of occupations in the real world. To ensure a fair system, researchers must make a \textit{normative decision} \cite{blodgett2020language} that they do not want to reproduce this particular correlation in their model.
This suggests that there may be inherent limitations to the ability of XAI methods to improve fairness of NLP methods through improving model robustness and generalization.

\paragraph{Some biases can be difficult for humans to recognize.} Even for biases that could be characterized in terms of surface-level correlations, XAI methods rely on humans to recognize what an undesirable correlation is, but biased models are often biased in subtle ways. For example, if the dialect bias in a hate speech detection system is mostly mediated by false positives on the uses of reclaimed slurs, this might seem like a good justification to a user who is unfamiliar with this phenomenon \citep{sap2019risk}. More studies with human subjects are needed to investigate whether humans can recognise unintended biases that cause fairness issues through explainability methods as well as they can recognise simpler data biases. 

\paragraph{Explainability methods are susceptible to fairwashing.} 
An issue that has repeatedly been raised with respect to XAI methods is the potential for ``fairwashing'' biased models. This refers to techniques that adversarially manipulate explanations in order to obscure the model's reliance on protected attributes. Fairwashing has been shown possible in rule lists \citep{aivodji2019fairwashing}, and both gradient based and perturbation based feature attribution methods \citep{dimanov2020you,anders2020fairwashing}. This relates to the wider issue of the faithfulness of an explainability method: if there is no guarantee that the explanations reflect the actual inner workings of the model, the explanations are of little use. One solution to this problem would be to extend certifiable robustness \citep{cohen2019certified, ma2021metamorphic} beyond the model itself, and develop certifiably faithful explainability methods with proofs that a particular way of testing for bias cannot be adversarially manipulated. Another approach to mitigate this issue is to provide the levels of uncertainty in the explanations, giving the end-user more information on whether to trust the generated explanation \citep{zhang2019should}, or other ways to calibrate user trust to the quality of the provided explanations \citep{zhang2020effect}. However, the effectiveness of these methods depends substantially 
on whether the model’s predicted probabilities are well-calibrated to the true outcome probabilities. Certain machine learning models do not meet this criterion. Specifically, the commonly used deep learning models have been shown to be over-confident in their predictions \cite{guo2017calibration}. Calibration of uncertainties is a necessary prerequisite, should they be used to calibrate user trust, as over-confident predictions can be themselves a source of mistrust. 

\paragraph{Fair AI is focused on outcome fairness, but XAI is motivated by procedural fairness.} Finally, it appears that there is a larger conceptual gap between the notions of fairness that the ethical AI community has developed, and the notion of fairness implicitly assumed in motivations for XAI methods. Namely, almost all the fairness metrics developed in Fair ML literature aim to formalize outcome fairness in that they are process-agnostic, and quantify the fairness of a model on its observed outcomes only. The type of fairness that motivates XAI, on the other hand, is closer to the concept of procedural fairness: XAI aims to elucidate the internal reasoning of a model, and make it transparent whether there are any parts of the decision process that could be deemed unfair. 

We observe that due to the lack of better definitions of procedural fairness, the most common way XAI methods are applied to fairness issues is to check whether the model uses features that are explicitly associated with protected attributes (e.g., gendered pronouns). This practice promotes a similar ideal with ``fairness through unawareness'' in that it aims to place the veil of ignorance about the protected attributes not at the level of the data fed into the model, but into the model itself. In other words, the best one could do with these techniques seem to be to develop ``colourblind'' models which, even if they receive explicit information about protected attributes in their input, ignore this information when making their decisions. Although it is simple and intuitive, we suspect that such an approach has similar issues with the much criticized ``fairness through unawareness'' approach \cite{kusner2017counterfactual, morse2021ends}. More clearly specified notions of procedural fairness, as well as precise quantitative metrics similar to those that have been developed for outcome fairness,  are needed in order to guide the development of XAI methods that can make ML models fairer. 



\section{Conclusion}

Publications in explainable NLP often cite \textit{fairness} as a motivation for the work, but the exact relationship between the two concepts is typically left unspecified. Most current XAI methods provide explanations on a local level through post-hoc processing, leaving open questions about how to automatically identify fairness issues in individual explanations, and how to generalize from local explanations to infer systematic model bias. Although the two fields of explainability and fairness feel intuitively linked, a review of the literature revealed a surprisingly small amount of work at the intersection. We have discussed some of the conceptual underpinnings shared by both these fields as well as practical challenges to uniting them, and proposed areas for future research. 

\bibliography{bibliography.bib}

\begin{thebibliography}{99}
\expandafter\ifx\csname natexlab\endcsname\relax\def\natexlab#1{#1}\fi

\bibitem[{Adadi and Berrada(2018)}]{adadi2018peeking}
Amina Adadi and Mohammed Berrada. 2018.
\newblock Peeking inside the black-box: {A} survey on explainable artificial
  intelligence {(XAI)}.
\newblock \emph{IEEE Access}, 6:52138--52160.

\bibitem[{Adragna et~al.(2020)Adragna, Creager, Madras, and
  Zemel}]{adragna2020fairness}
Robert Adragna, Elliot Creager, David Madras, and Richard Zemel. 2020.
\newblock Fairness and robustness in invariant learning: A case study in
  toxicity classification.
\newblock In \emph{Proceedings of the NeurIPS 2020 Workshop on Algorithmic
  Fairness through the Lens of Causality and Interpretability}.

\bibitem[{A{\"\i}vodji et~al.(2019)A{\"\i}vodji, Arai, Fortineau, Gambs, Hara,
  and Tapp}]{aivodji2019fairwashing}
Ulrich A{\"\i}vodji, Hiromi Arai, Olivier Fortineau, S{\'e}bastien Gambs,
  Satoshi Hara, and Alain Tapp. 2019.
\newblock Fairwashing: the risk of rationalization.
\newblock In \emph{Proceedings of the International Conference on Machine
  Learning}, pages 161--170.

\bibitem[{Aksenov et~al.(2021)Aksenov, Bourgonje, Zaczynska, Ostendorff,
  Schneider, and Rehm}]{aksenov2021fine}
Dmitrii Aksenov, Peter Bourgonje, Karolina Zaczynska, Malte Ostendorff,
  Julian~Moreno Schneider, and Georg Rehm. 2021.
\newblock Fine-grained classification of political bias in {G}erman news: A
  data set and initial experiments.
\newblock In \emph{Proceedings of the 5th Workshop on Online Abuse and Harms
  (WOAH 2021)}, pages 121--131.

\bibitem[{Alikhademi et~al.(2021)Alikhademi, Richardson, Drobina, and
  Gilbert}]{alikhademi2021can}
Kiana Alikhademi, Brianna Richardson, Emma Drobina, and Juan~E Gilbert. 2021.
\newblock Can explainable {AI} explain unfairness? {A} framework for evaluating
  explainable {AI}.
\newblock \emph{arXiv preprint arXiv:2106.07483}.

\bibitem[{Anders et~al.(2020)Anders, Pasliev, Dombrowski, M{\"u}ller, and
  Kessel}]{anders2020fairwashing}
Christopher Anders, Plamen Pasliev, Ann-Kathrin Dombrowski, Klaus-Robert
  M{\"u}ller, and Pan Kessel. 2020.
\newblock Fairwashing explanations with off-manifold detergent.
\newblock In \emph{Proceedings of the International Conference on Machine
  Learning}, pages 314--323.

\bibitem[{Bagdasaryan et~al.(2019)Bagdasaryan, Poursaeed, and
  Shmatikov}]{bagdasaryan2019differential}
Eugene Bagdasaryan, Omid Poursaeed, and Vitaly Shmatikov. 2019.
\newblock Differential privacy has disparate impact on model accuracy.
\newblock \emph{Advances in Neural Information Processing Systems}, 32.

\bibitem[{Bahdanau et~al.(2015)Bahdanau, Cho, and Bengio}]{bahdanau2015neural}
Dzmitry Bahdanau, Kyung~Hyun Cho, and Yoshua Bengio. 2015.
\newblock Neural machine translation by jointly learning to align and
  translate.
\newblock In \emph{Proceedings of the 3rd International Conference on Learning
  Representations}.

\bibitem[{Balashankar et~al.(2021)Balashankar, Wang, Packer, Thain, Chi, and
  Beutel}]{balashankar2021can}
Ananth Balashankar, Xuezhi Wang, Ben Packer, Nithum Thain, Ed~Chi, and Alex
  Beutel. 2021.
\newblock Can we improve model robustness through secondary attribute
  counterfactuals?
\newblock In \emph{Proceedings of the 2021 Conference on Empirical Methods in
  Natural Language Processing}, pages 4701--4712.

\bibitem[{Balk{\i}r et~al.(2022)Balk{\i}r, Nejadgholi, Fraser, and
  Kiritchenko}]{balkir2022necessity}
Esma Balk{\i}r, Isar Nejadgholi, Kathleen~C Fraser, and Svetlana Kiritchenko.
  2022.
\newblock Necessity and sufficiency for explaining text classifiers: A case
  study in hate speech detection.
\newblock In \emph{Proceedings of the Annual Conference of the North American
  Chapter of the Association for Computational Linguistics (NAACL)}, Seattle,
  WA, USA.

\bibitem[{Barocas and Selbst(2016)}]{barocas2016big}
Solon Barocas and Andrew~D Selbst. 2016.
\newblock Big data’s disparate impact.
\newblock \emph{CALIFORNIA LAW REVIEW}, 104:671.

\bibitem[{Begley et~al.(2020)Begley, Schwedes, Frye, and
  Feige}]{begley2020explainability}
Tom Begley, Tobias Schwedes, Christopher Frye, and Ilya Feige. 2020.
\newblock Explainability for fair machine learning.
\newblock \emph{arXiv preprint arXiv:2010.07389}.

\bibitem[{Blader and Tyler(2003)}]{blader2003constitutes}
Steven~L Blader and Tom~R Tyler. 2003.
\newblock What constitutes fairness in work settings? a four-component model of
  procedural justice.
\newblock \emph{Human Resource Management Review}, 13(1):107--126.

\bibitem[{Blodgett et~al.(2020)Blodgett, Barocas, Daum{\'e}~III, and
  Wallach}]{blodgett2020language}
Su~Lin Blodgett, Solon Barocas, Hal Daum{\'e}~III, and Hanna Wallach. 2020.
\newblock Language (technology) is power: A critical survey of “bias” in
  {NLP}.
\newblock In \emph{Proceedings of the 58th Annual Meeting of the Association
  for Computational Linguistics}, pages 5454--5476.

\bibitem[{Blodgett et~al.(2016)Blodgett, Green, and
  O’Connor}]{blodgett2016demographic}
Su~Lin Blodgett, Lisa Green, and Brendan O’Connor. 2016.
\newblock Demographic dialectal variation in social media: A case study of
  {A}frican-{A}merican {E}nglish.
\newblock In \emph{Proceedings of the 2016 Conference on Empirical Methods in
  Natural Language Processing}, pages 1119--1130.

\bibitem[{Blodgett and O'Connor(2017)}]{blodgett2017racial}
Su~Lin Blodgett and Brendan O'Connor. 2017.
\newblock Racial disparity in natural language processing: A case study of
  social media {A}frican-{A}merican {E}nglish.
\newblock In \emph{Proceedings of the 2017 Workshop on Fairness,
  Accountability, and Transparency in Machine Learning}.

\bibitem[{Bolukbasi et~al.(2016)Bolukbasi, Chang, Zou, Saligrama, and
  Kalai}]{bolukbasi2016man}
Tolga Bolukbasi, Kai-Wei Chang, James~Y Zou, Venkatesh Saligrama, and Adam~T
  Kalai. 2016.
\newblock Man is to computer programmer as woman is to homemaker? debiasing
  word embeddings.
\newblock \emph{Advances in Neural Information Processing Systems}, 29.

\bibitem[{Carlini et~al.(2021)Carlini, Tramer, Wallace, Jagielski,
  Herbert-Voss, Lee, Roberts, Brown, Song, Erlingsson, Oprea, and
  Raffel}]{carlini2021extracting}
Nicholas Carlini, Florian Tramer, Eric Wallace, Matthew Jagielski, Ariel
  Herbert-Voss, Katherine Lee, Adam Roberts, Tom Brown, Dawn Song, Ulfar
  Erlingsson, Alina Oprea, and Colin Raffel. 2021.
\newblock Extracting training data from large language models.
\newblock In \emph{Proceedings of the 30th USENIX Security Symposium}, pages
  2633--2650.

\bibitem[{Castelnovo et~al.(2022)Castelnovo, Crupi, Greco, Regoli, Penco, and
  Cosentini}]{castelnovo2021zoo}
Alessandro Castelnovo, Riccardo Crupi, Greta Greco, Daniele Regoli,
  Ilaria~Giuseppina Penco, and Andrea~Claudio Cosentini. 2022.
\newblock A clarification of the nuances in the fairness metrics landscape.
\newblock \emph{Scientific Reports}, 12.

\bibitem[{Caton and Haas(2020)}]{caton2020fairness}
Simon Caton and Christian Haas. 2020.
\newblock Fairness in machine learning: A survey.
\newblock \emph{arXiv preprint arXiv:2010.04053}.

\bibitem[{Chang et~al.(2019)Chang, Prabhakaran, and Ordonez}]{chang2019bias}
Kai-Wei Chang, Vinod Prabhakaran, and Vicente Ordonez. 2019.
\newblock Bias and fairness in natural language processing.
\newblock In \emph{Proceedings of the 2019 Conference on Empirical Methods in
  Natural Language Processing and the 9th International Joint Conference on
  Natural Language Processing (EMNLP-IJCNLP): Tutorial Abstracts}.

\bibitem[{Choi et~al.(2016)Choi, Bahadori, Sun, Kulas, Schuetz, and
  Stewart}]{choi2016retain}
Edward Choi, Mohammad~Taha Bahadori, Jimeng Sun, Joshua Kulas, Andy Schuetz,
  and Walter Stewart. 2016.
\newblock Retain: An interpretable predictive model for healthcare using
  reverse time attention mechanism.
\newblock \emph{Advances in Neural Information Processing Systems}, 29.

\bibitem[{Clark et~al.(2019)Clark, Khandelwal, Levy, and
  Manning}]{clark2019does}
Kevin Clark, Urvashi Khandelwal, Omer Levy, and Christopher~D Manning. 2019.
\newblock What does {BERT} look at? {A}n analysis of {BERT}’s attention.
\newblock In \emph{Proceedings of the 2019 ACL Workshop BlackboxNLP: Analyzing
  and Interpreting Neural Networks for NLP}, pages 276--286.

\bibitem[{Cohen et~al.(2019)Cohen, Rosenfeld, and Kolter}]{cohen2019certified}
Jeremy Cohen, Elan Rosenfeld, and Zico Kolter. 2019.
\newblock Certified adversarial robustness via randomized smoothing.
\newblock In \emph{Proceedings of the International Conference on Machine
  Learning}, pages 1310--1320.

\bibitem[{Czarnowska et~al.(2021)Czarnowska, Vyas, and
  Shah}]{czarnowska2021quantifying}
Paula Czarnowska, Yogarshi Vyas, and Kashif Shah. 2021.
\newblock Quantifying social biases in {NLP}: A generalization and empirical
  comparison of extrinsic fairness metrics.
\newblock \emph{Transactions of the Association for Computational Linguistics},
  9:1249--1267.

\bibitem[{Danilevsky et~al.(2020)Danilevsky, Qian, Aharonov, Katsis, Kawas, and
  Sen}]{danilevsky2020survey}
Marina Danilevsky, Kun Qian, Ranit Aharonov, Yannis Katsis, Ban Kawas, and
  Prithviraj Sen. 2020.
\newblock \href {https://aclanthology.org/2020.aacl-main.46} {A survey of the
  state of explainable {AI} for natural language processing}.
\newblock In \emph{Proceedings of the 1st Conference of the Asia-Pacific
  Chapter of the Association for Computational Linguistics and the 10th
  International Joint Conference on Natural Language Processing}, pages
  447--459, Suzhou, China. Association for Computational Linguistics.

\bibitem[{Das and Rad(2020)}]{das2020opportunities}
Arun Das and Paul Rad. 2020.
\newblock Opportunities and challenges in explainable artificial intelligence
  (xai): A survey.
\newblock \emph{arXiv preprint arXiv:2006.11371}.

\bibitem[{DeYoung et~al.(2020)DeYoung, Jain, Rajani, Lehman, Xiong, Socher, and
  Wallace}]{deyoung2020eraser}
Jay DeYoung, Sarthak Jain, Nazneen~Fatema Rajani, Eric Lehman, Caiming Xiong,
  Richard Socher, and Byron~C Wallace. 2020.
\newblock Eraser: A benchmark to evaluate rationalized {NLP} models.
\newblock In \emph{Proceedings of the 58th Annual Meeting of the Association
  for Computational Linguistics}, pages 4443--4458.

\bibitem[{Dimanov et~al.(2020)Dimanov, Bhatt, Jamnik, and
  Weller}]{dimanov2020you}
Botty Dimanov, Umang Bhatt, Mateja Jamnik, and Adrian Weller. 2020.
\newblock You shouldn't trust me: Learning models which conceal unfairness from
  multiple explanation methods.
\newblock In \emph{Proceedings of the AAAI Workshop on Artificial Intelligence
  Safety (SafeAI)}.

\bibitem[{Doshi-Velez and Kim(2017)}]{doshi2017towards}
Finale Doshi-Velez and Been Kim. 2017.
\newblock Towards a rigorous science of interpretable machine learning.
\newblock \emph{arXiv preprint arXiv:1702.08608}.

\bibitem[{Feder et~al.(2021{\natexlab{a}})Feder, Keith, Manzoor, Pryzant,
  Sridhar, Wood-Doughty, Eisenstein, Grimmer, Reichart, Roberts
  et~al.}]{feder2021causal}
Amir Feder, Katherine~A Keith, Emaad Manzoor, Reid Pryzant, Dhanya Sridhar,
  Zach Wood-Doughty, Jacob Eisenstein, Justin Grimmer, Roi Reichart, Margaret~E
  Roberts, et~al. 2021{\natexlab{a}}.
\newblock Causal inference in natural language processing: Estimation,
  prediction, interpretation and beyond.
\newblock \emph{arXiv preprint arXiv:2109.00725}.

\bibitem[{Feder et~al.(2021{\natexlab{b}})Feder, Oved, Shalit, and
  Reichart}]{feder2021causalm}
Amir Feder, Nadav Oved, Uri Shalit, and Roi Reichart. 2021{\natexlab{b}}.
\newblock Causalm: Causal model explanation through counterfactual language
  models.
\newblock \emph{Computational Linguistics}, 47(2):333--386.

\bibitem[{Galassi et~al.(2020)Galassi, Lippi, and
  Torroni}]{galassi2020attention}
Andrea Galassi, Marco Lippi, and Paolo Torroni. 2020.
\newblock Attention in natural language processing.
\newblock \emph{IEEE Transactions on Neural Networks and Learning Systems},
  32(10):4291--4308.

\bibitem[{Gardner et~al.(2020)Gardner, Artzi, Basmov, Berant, Bogin, Chen,
  Dasigi, Dua, Elazar, Gottumukkala et~al.}]{gardner2020evaluating}
Matt Gardner, Yoav Artzi, Victoria Basmov, Jonathan Berant, Ben Bogin, Sihao
  Chen, Pradeep Dasigi, Dheeru Dua, Yanai Elazar, Ananth Gottumukkala, et~al.
  2020.
\newblock Evaluating models’ local decision boundaries via contrast sets.
\newblock In \emph{Findings of the Association for Computational Linguistics:
  EMNLP 2020}, pages 1307--1323.

\bibitem[{Garg et~al.(2019)Garg, Perot, Limtiaco, Taly, Chi, and
  Beutel}]{garg2019counterfactual}
Sahaj Garg, Vincent Perot, Nicole Limtiaco, Ankur Taly, Ed~H Chi, and Alex
  Beutel. 2019.
\newblock Counterfactual fairness in text classification through robustness.
\newblock In \emph{Proceedings of the 2019 AAAI/ACM Conference on AI, Ethics,
  and Society}, pages 219--226.

\bibitem[{Goldfarb-Tarrant et~al.(2021)Goldfarb-Tarrant, Marchant, S{\'a}nchez,
  Pandya, and Lopez}]{goldfarb2021intrinsic}
Seraphina Goldfarb-Tarrant, Rebecca Marchant, Ricardo~Mu{\~n}oz S{\'a}nchez,
  Mugdha Pandya, and Adam Lopez. 2021.
\newblock Intrinsic bias metrics do not correlate with application bias.
\newblock In \emph{Proceedings of the 59th Annual Meeting of the Association
  for Computational Linguistics and the 11th International Joint Conference on
  Natural Language Processing (Volume 1: Long Papers)}, pages 1926--1940.

\bibitem[{Grabowicz et~al.(2022)Grabowicz, Perello, and
  Mishra}]{grabowicz2022marrying}
Przemyslaw Grabowicz, Nicholas Perello, and Aarshee Mishra. 2022.
\newblock Marrying fairness and explainability in supervised learning.
\newblock \emph{arXiv preprint arXiv:2204.02947}.

\bibitem[{Grgi{\'c}-Hla{\v{c}}a et~al.(2018)Grgi{\'c}-Hla{\v{c}}a, Zafar,
  Gummadi, and Weller}]{grgic2018beyond}
Nina Grgi{\'c}-Hla{\v{c}}a, Muhammad~Bilal Zafar, Krishna~P Gummadi, and Adrian
  Weller. 2018.
\newblock Beyond distributive fairness in algorithmic decision making: Feature
  selection for procedurally fair learning.
\newblock In \emph{Proceedings of the AAAI Conference on Artificial
  Intelligence}, volume~32.

\bibitem[{Guidotti et~al.(2018)Guidotti, Monreale, Ruggieri, Turini, Giannotti,
  and Pedreschi}]{Guidotti2018}
Riccardo Guidotti, Anna Monreale, Salvatore Ruggieri, Franco Turini, Fosca
  Giannotti, and Dino Pedreschi. 2018.
\newblock A survey of methods for explaining black box models.
\newblock \emph{ACM Computing Surveys (CSUR)}, 51(5):1--42.

\bibitem[{Guo et~al.(2017)Guo, Pleiss, Sun, and
  Weinberger}]{guo2017calibration}
Chuan Guo, Geoff Pleiss, Yu~Sun, and Kilian~Q Weinberger. 2017.
\newblock On calibration of modern neural networks.
\newblock In \emph{Proceedings of the International Conference on Machine
  Learning}, pages 1321--1330.

\bibitem[{Han et~al.(2020)Han, Wallace, and Tsvetkov}]{han2020explaining}
Xiaochuang Han, Byron~C Wallace, and Yulia Tsvetkov. 2020.
\newblock Explaining black box predictions and unveiling data artifacts through
  influence functions.
\newblock In \emph{Proceedings of the 58th Annual Meeting of the Association
  for Computational Linguistics}, pages 5553--5563.

\bibitem[{Hewitt and Liang(2019)}]{hewitt2019designing}
John Hewitt and Percy Liang. 2019.
\newblock Designing and interpreting probes with control tasks.
\newblock In \emph{Proceedings of the 2019 Conference on Empirical Methods in
  Natural Language Processing and the 9th International Joint Conference on
  Natural Language Processing (EMNLP-IJCNLP)}, pages 2733--2743.

\bibitem[{Hooker(2021)}]{hooker2021moving}
Sara Hooker. 2021.
\newblock Moving beyond “algorithmic bias is a data problem”.
\newblock \emph{Patterns}, 2(4):100241.

\bibitem[{Jain and Wallace(2019)}]{jain2019attention}
Sarthak Jain and Byron~C Wallace. 2019.
\newblock Attention is not explanation.
\newblock In \emph{Proceedings of the 2019 Conference of the North American
  Chapter of the Association for Computational Linguistics: Human Language
  Technologies, Volume 1 (Long and Short Papers)}, pages 3543--3556.

\bibitem[{Jin et~al.(2019)Jin, Wei, Du, Xue, and Ren}]{jin2019towards}
Xisen Jin, Zhongyu Wei, Junyi Du, Xiangyang Xue, and Xiang Ren. 2019.
\newblock Towards hierarchical importance attribution: Explaining compositional
  semantics for neural sequence models.
\newblock In \emph{Proceedings of the International Conference on Learning
  Representations}.

\bibitem[{Kaushik et~al.(2020)Kaushik, Setlur, Hovy, and
  Lipton}]{kaushik2020explaining}
Divyansh Kaushik, Amrith Setlur, Eduard~H Hovy, and Zachary~Chase Lipton. 2020.
\newblock Explaining the efficacy of counterfactually augmented data.
\newblock In \emph{Proceedings of the International Conference on Learning
  Representations}.

\bibitem[{Kennedy et~al.(2020)Kennedy, Jin, Davani, Dehghani, and
  Ren}]{kennedy2020contextualizing}
Brendan Kennedy, Xisen Jin, Aida~Mostafazadeh Davani, Morteza Dehghani, and
  Xiang Ren. 2020.
\newblock Contextualizing hate speech classifiers with post-hoc explanation.
\newblock In \emph{Proceedings of the 58th Annual Meeting of the Association
  for Computational Linguistics}, pages 5435--5442.

\bibitem[{Kim et~al.(2018)Kim, Wattenberg, Gilmer, Cai, Wexler, Viegas
  et~al.}]{kim2018interpretability}
Been Kim, Martin Wattenberg, Justin Gilmer, Carrie Cai, James Wexler, Fernanda
  Viegas, et~al. 2018.
\newblock Interpretability beyond feature attribution: Quantitative testing
  with concept activation vectors (tcav).
\newblock In \emph{Proceedings of the International Conference on Machine
  Learning}, pages 2668--2677.

\bibitem[{Kiritchenko and Mohammad(2018)}]{kiritchenko2018examining}
Svetlana Kiritchenko and Saif Mohammad. 2018.
\newblock Examining gender and race bias in two hundred sentiment analysis
  systems.
\newblock In \emph{Proceedings of the Seventh Joint Conference on Lexical and
  Computational Semantics}, pages 43--53.

\bibitem[{Koh and Liang(2017)}]{koh2017understanding}
Pang~Wei Koh and Percy Liang. 2017.
\newblock Understanding black-box predictions via influence functions.
\newblock In \emph{Proceedings of the International Conference on Machine
  Learning}, pages 1885--1894.

\bibitem[{Kusner et~al.(2017)Kusner, Loftus, Russell, and
  Silva}]{kusner2017counterfactual}
Matt~J Kusner, Joshua Loftus, Chris Russell, and Ricardo Silva. 2017.
\newblock Counterfactual fairness.
\newblock \emph{Advances in neural information processing systems}, 30.

\bibitem[{Lertvittayakumjorn and
  Toni(2021)}]{lertvittayakumjorn2021explanation}
Piyawat Lertvittayakumjorn and Francesca Toni. 2021.
\newblock Explanation-based human debugging of {NLP} models: A survey.
\newblock \emph{Transactions of the Association for Computational Linguistics},
  9:1508--1528.

\bibitem[{Loftus et~al.(2018)Loftus, Russell, Kusner, and
  Silva}]{loftus2018causal}
Joshua~R Loftus, Chris Russell, Matt~J Kusner, and Ricardo Silva. 2018.
\newblock Causal reasoning for algorithmic fairness.
\newblock \emph{arXiv preprint arXiv:1805.05859}.

\bibitem[{Lundberg and Lee(2017)}]{lundberg2017A}
Scott~M Lundberg and Su-In Lee. 2017.
\newblock \href
  {http://papers.nips.cc/paper/7062-a-unified-approach-to-interpreting-model-predictions.pdf}
  {A unified approach to interpreting model predictions}.
\newblock In I.~Guyon, U.~V. Luxburg, S.~Bengio, H.~Wallach, R.~Fergus,
  S.~Vishwanathan, and R.~Garnett, editors, \emph{Advances in Neural
  Information Processing Systems 30}, pages 4765--4774. Curran Associates, Inc.

\bibitem[{Ma et~al.(2021)Ma, Wang, and Liu}]{ma2021metamorphic}
Pingchuan Ma, Shuai Wang, and Jin Liu. 2021.
\newblock Metamorphic testing and certified mitigation of fairness violations
  in {NLP} models.
\newblock In \emph{Proceedings of the Twenty-Ninth International Joint
  Conferences on Artificial Intelligence}, pages 458--465.

\bibitem[{Mathew et~al.(2021)Mathew, Saha, Yimam, Biemann, Goyal, and
  Mukherjee}]{mathew2021hatexplain}
Binny Mathew, Punyajoy Saha, Seid~Muhie Yimam, Chris Biemann, Pawan Goyal, and
  Animesh Mukherjee. 2021.
\newblock Hatexplain: A benchmark dataset for explainable hate speech
  detection.
\newblock In \emph{Proceedings of the AAAI Conference on Artificial
  Intelligence}, volume~35, pages 14867--14875.

\bibitem[{Mehrabi et~al.(2021)Mehrabi, Morstatter, Saxena, Lerman, and
  Galstyan}]{mehrabi2021survey}
Ninareh Mehrabi, Fred Morstatter, Nripsuta Saxena, Kristina Lerman, and Aram
  Galstyan. 2021.
\newblock A survey on bias and fairness in machine learning.
\newblock \emph{ACM Computing Surveys (CSUR)}, 54(6):1--35.

\bibitem[{Mohammad(2022)}]{mohammad2021ethics}
Saif~M Mohammad. 2022.
\newblock Ethics sheet for automatic emotion recognition and sentiment
  analysis.
\newblock \emph{Computational Linguistics}.

\bibitem[{Morley et~al.(2021)Morley, Floridi, Kinsey, and
  Elhalal}]{morley2021initial}
Jessica Morley, Luciano Floridi, Libby Kinsey, and Anat Elhalal. 2021.
\newblock From what to how: an initial review of publicly available {AI} ethics
  tools, methods and research to translate principles into practices.
\newblock In \emph{Ethics, Governance, and Policies in Artificial
  Intelligence}, pages 153--183. Springer.

\bibitem[{Morse et~al.(2021)Morse, Teodorescu, Awwad, and Kane}]{morse2021ends}
Lily Morse, Mike Horia~M Teodorescu, Yazeed Awwad, and Gerald~C Kane. 2021.
\newblock Do the ends justify the means? variation in the distributive and
  procedural fairness of machine learning algorithms.
\newblock \emph{Journal of Business Ethics}, pages 1--13.

\bibitem[{Mosca et~al.(2021)Mosca, Wich, and Groh}]{mosca2021understanding}
Edoardo Mosca, Maximilian Wich, and Georg Groh. 2021.
\newblock Understanding and interpreting the impact of user context in hate
  speech detection.
\newblock In \emph{Proceedings of the Ninth International Workshop on Natural
  Language Processing for Social Media}, pages 91--102.

\bibitem[{Nejadgholi et~al.(2022)Nejadgholi, Fraser, and
  Kiritchenko}]{nejadgholi2022improving}
Isar Nejadgholi, Kathleen~C Fraser, and Svetlana Kiritchenko. 2022.
\newblock Improving generalizability in implicitly abusive language detection
  with concept activation vectors.
\newblock In \emph{Proceedings of the 60th Annual Meeting of the Association
  for Computational Linguistics}, Dublin, Ireland.

\bibitem[{P et~al.(2021)P, V, and Jose}]{deepak2021fairness}
Deepak P, Sanil V, and Joemon~M. Jose. 2021.
\newblock On fairness and interpretability.
\newblock In \emph{Proceedings of the IJCAI Workshop on AI for Social Good}.

\bibitem[{Parafita and Vitria(2021)}]{parafita2021deep}
Alvaro Parafita and Jordi Vitria. 2021.
\newblock Deep causal graphs for causal inference, black-box explainability and
  fairness.
\newblock In \emph{Artificial Intelligence Research and Development:
  Proceedings of the 23rd International Conference of the Catalan Association
  for Artificial Intelligence}, volume 339, page 415. IOS Press.

\bibitem[{Pearl(2009)}]{pearl2009causality}
Judea Pearl. 2009.
\newblock \emph{Causality}.
\newblock Cambridge University Press.

\bibitem[{Prabhakaran et~al.(2019)Prabhakaran, Hutchinson, and
  Mitchell}]{prabhakaran2019perturbation}
Vinodkumar Prabhakaran, Ben Hutchinson, and Margaret Mitchell. 2019.
\newblock Perturbation sensitivity analysis to detect unintended model biases.
\newblock In \emph{Proceedings of the 2019 Conference on Empirical Methods in
  Natural Language Processing and the 9th International Joint Conference on
  Natural Language Processing (EMNLP-IJCNLP)}, pages 5740--5745.

\bibitem[{Pradhan et~al.(2022)Pradhan, Zhu, Glavic, and
  Salimi}]{pradhan2021interpretable}
Romila Pradhan, Jiongli Zhu, Boris Glavic, and Babak Salimi. 2022.
\newblock Interpretable data-based explanations for fairness debugging.
\newblock In \emph{Proceedings of the 2022 ACM SIGMOD International Conference
  on Management of Data}.

\bibitem[{Pruksachatkun et~al.(2021)Pruksachatkun, Krishna, Dhamala, Gupta, and
  Chang}]{pruksachatkun2021does}
Yada Pruksachatkun, Satyapriya Krishna, Jwala Dhamala, Rahul Gupta, and Kai-Wei
  Chang. 2021.
\newblock Does robustness improve fairness? approaching fairness with word
  substitution robustness methods for text classification.
\newblock In \emph{Findings of the Association for Computational Linguistics:
  ACL-IJCNLP 2021}, pages 3320--3331.

\bibitem[{Ribeiro et~al.(2016)Ribeiro, Singh, and Guestrin}]{ribeiro2016should}
Marco~Tulio Ribeiro, Sameer Singh, and Carlos Guestrin. 2016.
\newblock ``{W}hy should i trust you?'' {E}xplaining the predictions of any
  classifier.
\newblock In \emph{Proceedings of the 22nd ACM SIGKDD International Conference
  on Knowledge Discovery and Data Mining}, pages 1135--1144.

\bibitem[{Ribeiro et~al.(2018{\natexlab{a}})Ribeiro, Singh, and
  Guestrin}]{ribeiro2018anchors}
Marco~Tulio Ribeiro, Sameer Singh, and Carlos Guestrin. 2018{\natexlab{a}}.
\newblock Anchors: High-precision model-agnostic explanations.
\newblock In \emph{Proceedings of the AAAI Conference on Artificial
  Intelligence}, volume~32.

\bibitem[{Ribeiro et~al.(2018{\natexlab{b}})Ribeiro, Singh, and
  Guestrin}]{ribeiro2018semantically}
Marco~Tulio Ribeiro, Sameer Singh, and Carlos Guestrin. 2018{\natexlab{b}}.
\newblock Semantically equivalent adversarial rules for debugging {NLP} models.
\newblock In \emph{Proceedings of the 56th Annual Meeting of the Association
  for Computational Linguistics (Volume 1: Long Papers)}, pages 856--865.

\bibitem[{Ribeiro et~al.(2020)Ribeiro, Wu, Guestrin, and
  Singh}]{ribeiro2020beyond}
Marco~Tulio Ribeiro, Tongshuang Wu, Carlos Guestrin, and Sameer Singh. 2020.
\newblock Beyond accuracy: Behavioral testing of nlp models with checklist.
\newblock In \emph{Proceedings of the 58th Annual Meeting of the Association
  for Computational Linguistics}, pages 4902--4912.

\bibitem[{Ross et~al.(2021)Ross, Marasovi{\'c}, and
  Peters}]{ross2021explaining}
Alexis Ross, Ana Marasovi{\'c}, and Matthew~E Peters. 2021.
\newblock Explaining {NLP} models via minimal contrastive editing ({MiCE}).
\newblock In \emph{Findings of the Association for Computational Linguistics:
  ACL-IJCNLP 2021}, pages 3840--3852.

\bibitem[{R{\"o}ttger et~al.(2021)R{\"o}ttger, Vidgen, Nguyen, Waseem,
  Margetts, and Pierrehumbert}]{rottger2021hatecheck}
Paul R{\"o}ttger, Bertie Vidgen, Dong Nguyen, Zeerak Waseem, Helen Margetts,
  and Janet Pierrehumbert. 2021.
\newblock Hatecheck: Functional tests for hate speech detection models.
\newblock In \emph{Proceedings of the 59th Annual Meeting of the Association
  for Computational Linguistics and the 11th International Joint Conference on
  Natural Language Processing (Volume 1: Long Papers)}, pages 41--58.

\bibitem[{Rudinger et~al.(2018)Rudinger, Naradowsky, Leonard, and
  Van~Durme}]{rudinger2018gender}
Rachel Rudinger, Jason Naradowsky, Brian Leonard, and Benjamin Van~Durme. 2018.
\newblock Gender bias in coreference resolution.
\newblock In \emph{Proceedings of the 2018 Conference of the North American
  Chapter of the Association for Computational Linguistics: Human Language
  Technologies, Volume 2 (Short Papers)}, pages 8--14.

\bibitem[{Sap et~al.(2019)Sap, Card, Gabriel, Choi, and Smith}]{sap2019risk}
Maarten Sap, Dallas Card, Saadia Gabriel, Yejin Choi, and Noah~A Smith. 2019.
\newblock The risk of racial bias in hate speech detection.
\newblock In \emph{Proceedings of the 57th Annual Meeting of the Association
  for Computational Linguistics}, pages 1668--1678.

\bibitem[{Selvaraju et~al.(2017)Selvaraju, Cogswell, Das, Vedantam, Parikh, and
  Batra}]{selvaraju2017grad}
Ramprasaath~R Selvaraju, Michael Cogswell, Abhishek Das, Ramakrishna Vedantam,
  Devi Parikh, and Dhruv Batra. 2017.
\newblock Grad-cam: Visual explanations from deep networks via gradient-based
  localization.
\newblock In \emph{Proceedings of the IEEE International Conference on Computer
  Vision}, pages 618--626.

\bibitem[{Shah et~al.(2020)Shah, Schwartz, and Hovy}]{shah2020predictive}
Deven~Santosh Shah, H~Andrew Schwartz, and Dirk Hovy. 2020.
\newblock Predictive biases in natural language processing models: A conceptual
  framework and overview.
\newblock In \emph{Proceedings of the 58th Annual Meeting of the Association
  for Computational Linguistics}, pages 5248--5264.

\bibitem[{Shrikumar et~al.(2017)Shrikumar, Greenside, and
  Kundaje}]{shrikumar2017learning}
Avanti Shrikumar, Peyton Greenside, and Anshul Kundaje. 2017.
\newblock Learning important features through propagating activation
  differences.
\newblock In \emph{Proceedings of the International Conference on Machine
  Learning}, pages 3145--3153.

\bibitem[{Simonyan et~al.(2014)Simonyan, Vedaldi, and
  Zisserman}]{simonyan2013deep}
Karen Simonyan, Andrea Vedaldi, and Andrew Zisserman. 2014.
\newblock Deep inside convolutional networks: Visualising image classification
  models and saliency maps.
\newblock In \emph{Proceedings of the 2nd International Conference on Learning
  Representations, Workshop Track}.

\bibitem[{Smilkov et~al.(2017)Smilkov, Thorat, Kim, Vi{\'e}gas, and
  Wattenberg}]{smilkov2017smoothgrad}
Daniel Smilkov, Nikhil Thorat, Been Kim, Fernanda Vi{\'e}gas, and Martin
  Wattenberg. 2017.
\newblock Smoothgrad: removing noise by adding noise.
\newblock \emph{arXiv preprint arXiv:1706.03825}.

\bibitem[{Stanovsky et~al.(2019)Stanovsky, Smith, and
  Zettlemoyer}]{stanovsky2019evaluating}
Gabriel Stanovsky, Noah~A Smith, and Luke Zettlemoyer. 2019.
\newblock Evaluating gender bias in machine translation.
\newblock In \emph{Proceedings of the 57th Annual Meeting of the Association
  for Computational Linguistics}, pages 1679--1684.

\bibitem[{Sundararajan et~al.(2017)Sundararajan, Taly, and
  Yan}]{sundararajan2017axiomatic}
Mukund Sundararajan, Ankur Taly, and Qiqi Yan. 2017.
\newblock Axiomatic attribution for deep networks.
\newblock In \emph{Proceedings of the International Conference on Machine
  Learning}, pages 3319--3328.

\bibitem[{Tan et~al.(2018)Tan, Caruana, Hooker, and Lou}]{tan2018distill}
Sarah Tan, Rich Caruana, Giles Hooker, and Yin Lou. 2018.
\newblock Distill-and-compare: Auditing black-box models using transparent
  model distillation.
\newblock In \emph{Proceedings of the 2018 AAAI/ACM Conference on AI, Ethics,
  and Society}, pages 303--310.

\bibitem[{Voita and Titov(2020)}]{voita2020information}
Elena Voita and Ivan Titov. 2020.
\newblock Information-theoretic probing with minimum description length.
\newblock In \emph{Proceedings of the 2020 Conference on Empirical Methods in
  Natural Language Processing (EMNLP)}, pages 183--196.

\bibitem[{Wallace et~al.(2020)Wallace, Gardner, and
  Singh}]{wallace2020interpreting}
Eric Wallace, Matt Gardner, and Sameer Singh. 2020.
\newblock Interpreting predictions of {NLP} models.
\newblock In \emph{Proceedings of the 2020 Conference on Empirical Methods in
  Natural Language Processing: Tutorial Abstracts}, pages 20--23.

\bibitem[{Wallace et~al.(2019)Wallace, Tuyls, Wang, Subramanian, Gardner, and
  Singh}]{wallace2019allennlp}
Eric Wallace, Jens Tuyls, Junlin Wang, Sanjay Subramanian, Matt Gardner, and
  Sameer Singh. 2019.
\newblock {AllenNLP} {I}nterpret: A framework for explaining predictions of
  {NLP} models.
\newblock In \emph{Proceedings of the 2019 Conference on Empirical Methods in
  Natural Language Processing and the 9th International Joint Conference on
  Natural Language Processing (EMNLP-IJCNLP): System Demonstrations}, pages
  7--12.

\bibitem[{Wich et~al.(2020)Wich, Bauer, and Groh}]{wich2020impact}
Maximilian Wich, Jan Bauer, and Georg Groh. 2020.
\newblock Impact of politically biased data on hate speech classification.
\newblock In \emph{Proceedings of the Fourth Workshop on Online Abuse and
  Harms}, pages 54--64.

\bibitem[{Wiegreffe and Pinter(2019)}]{wiegreffe2019attention}
Sarah Wiegreffe and Yuval Pinter. 2019.
\newblock Attention is not not explanation.
\newblock In \emph{Proceedings of the 2019 Conference on Empirical Methods in
  Natural Language Processing and the 9th International Joint Conference on
  Natural Language Processing (EMNLP-IJCNLP)}, pages 11--20.

\bibitem[{Wu et~al.(2021)Wu, Ribeiro, Heer, and Weld}]{wu2021polyjuice}
Tongshuang Wu, Marco~Tulio Ribeiro, Jeffrey Heer, and Daniel Weld. 2021.
\newblock \href {https://doi.org/10.18653/v1/2021.acl-long.523} {Polyjuice:
  Generating counterfactuals for explaining, evaluating, and improving models}.
\newblock In \emph{Proceedings of the 59th Annual Meeting of the Association
  for Computational Linguistics and the 11th International Joint Conference on
  Natural Language Processing (Volume 1: Long Papers)}, pages 6707--6723,
  Online. Association for Computational Linguistics.

\bibitem[{Xu et~al.(2015)Xu, Ba, Kiros, Cho, Courville, Salakhudinov, Zemel,
  and Bengio}]{xu2015show}
Kelvin Xu, Jimmy Ba, Ryan Kiros, Kyunghyun Cho, Aaron Courville, Ruslan
  Salakhudinov, Rich Zemel, and Yoshua Bengio. 2015.
\newblock Show, attend and tell: Neural image caption generation with visual
  attention.
\newblock In \emph{Proceedings of the International Conference on Machine
  Learning}, pages 2048--2057.

\bibitem[{Yang et~al.(2021)Yang, Li, Cunningham, Zhang, Smyth, and
  Dong}]{yang2021exploring}
Linyi Yang, Jiazheng Li, P{\'a}draig Cunningham, Yue Zhang, Barry Smyth, and
  Ruihai Dong. 2021.
\newblock Exploring the efficacy of automatically generated counterfactuals for
  sentiment analysis.
\newblock In \emph{Proceedings of the 59th Annual Meeting of the Association
  for Computational Linguistics and the 11th International Joint Conference on
  Natural Language Processing (Volume 1: Long Papers)}, pages 306--316.

\bibitem[{Yeh et~al.(2018)Yeh, Kim, Yen, and Ravikumar}]{yeh2018representer}
Chih-Kuan Yeh, Joon Kim, Ian En-Hsu Yen, and Pradeep~K Ravikumar. 2018.
\newblock Representer point selection for explaining deep neural networks.
\newblock \emph{Advances in Neural Information Processing Systems}, 31.

\bibitem[{Zhang and Bareinboim(2018)}]{zhang2018fairness}
Junzhe Zhang and Elias Bareinboim. 2018.
\newblock Fairness in decision-making—the causal explanation formula.
\newblock In \emph{Proceedings of the AAAI Conference on Artificial
  Intelligence}, volume~32.

\bibitem[{Zhang et~al.(2020{\natexlab{a}})Zhang, Sheng, Alhazmi, and
  Li}]{zhang2020adversarial}
Wei~Emma Zhang, Quan~Z Sheng, Ahoud Alhazmi, and Chenliang Li.
  2020{\natexlab{a}}.
\newblock Adversarial attacks on deep-learning models in natural language
  processing: A survey.
\newblock \emph{ACM Transactions on Intelligent Systems and Technology (TIST)},
  11(3):1--41.

\bibitem[{Zhang et~al.(2019)Zhang, Song, Sun, Tan, and Udell}]{zhang2019should}
Yujia Zhang, Kuangyan Song, Yiming Sun, Sarah Tan, and Madeleine Udell. 2019.
\newblock "why should you trust my explanation?" understanding uncertainty in
  {LIME} explanations.
\newblock In \emph{Proceedings of the ICML Workshop AI for Social Good}.

\bibitem[{Zhang et~al.(2020{\natexlab{b}})Zhang, Liao, and
  Bellamy}]{zhang2020effect}
Yunfeng Zhang, Q~Vera Liao, and Rachel~KE Bellamy. 2020{\natexlab{b}}.
\newblock Effect of confidence and explanation on accuracy and trust
  calibration in {AI}-assisted decision making.
\newblock In \emph{Proceedings of the 2020 Conference on Fairness,
  Accountability, and Transparency}, pages 295--305.

\bibitem[{Zhao et~al.(2019)Zhao, Wang, Yatskar, Cotterell, Ordonez, and
  Chang}]{zhao2019gender}
Jieyu Zhao, Tianlu Wang, Mark Yatskar, Ryan Cotterell, Vicente Ordonez, and
  Kai-Wei Chang. 2019.
\newblock Gender bias in contextualized word embeddings.
\newblock In \emph{Proceedings of the 2019 Conference of the North American
  Chapter of the Association for Computational Linguistics: Human Language
  Technologies}, volume~1.

\bibitem[{Zylberajch et~al.(2021)Zylberajch, Lertvittayakumjorn, and
  Toni}]{zylberajch2021hildif}
Hugo Zylberajch, Piyawat Lertvittayakumjorn, and Francesca Toni. 2021.
\newblock Hildif: Interactive debugging of {NLI} models using influence
  functions.
\newblock In \emph{Proceedings of the First Workshop on Interactive Learning
  for Natural Language Processing}, pages 1--6.

\end{thebibliography}
\bibliographystyle{acl_natbib}

\end{document}